\documentclass[conference]{IEEEtran}
\usepackage{cite}
\usepackage{amsmath,amssymb,amsfonts}
\usepackage{algorithmic}
\usepackage{graphicx}
\usepackage{subfig}
\usepackage{textcomp}
\usepackage{xcolor}
\usepackage{listings}
\usepackage{caption} 
\usepackage{authblk}

\def\BibTeX{{\rm B\kern-.05em{\sc i\kern-.025em b}\kern-.08em
    T\kern-.1667em\lower.7ex\hbox{E}\kern-.125emX}}
\begin{document}

\title{Single Storage Semi-Global Matching for Real Time Depth Processing\\
}

\author[1]{Prathmesh Sawant}
\author[1]{Yashwant Temburu}
\author[1]{Mandar Datar}
\author[2]{Imran Ahmed}
\author[2]{Vinayak Shriniwas}
\author[1]{Sachin Patkar}
\affil[1]{Department of Electrical Engineering, Indian Institute of Technology Bombay, India }
\affil[2]{Defence Research and Development Organisation, India}

\renewcommand\Authands{ and }

\maketitle

\begin{abstract}
Depth-map is the key computation in computer vision and robotics. One of the  most popular approach is via computation of disparity-map of images obtained from Stereo Camera.  Semi Global Matching (SGM) method is a popular choice for good accuracy with reasonable computation time. To use such compute-intensive algorithms for real-time applications such as for autonomous aerial vehicles, blind Aid, etc. acceleration using GPU, FPGA is necessary. In this paper, we show the design and implementation of a stereo-vision system, which is based on FPGA-implementation of More Global Matching(MGM)\cite{mgm}. MGM is a variant of SGM. We use 4 paths but store a single cumulative cost value for a corresponding pixel. Our stereo-vision prototype uses Zedboard containing an ARM-based Zynq-SoC \cite{zedboard}, ZED-stereo-camera / ELP stereo-camera / Intel RealSense D435i, and VGA for visualization. 
The power consumption attributed to the custom FPGA-based acceleration of disparity map computation required for depth-map is just 0.72 watt. The update rate of the disparity map is realistic 10.5 fps. 
\end{abstract}


\section{Introduction}
Although 2D and 3D LIDARs (Light Detection and Ranging Sensors) provided accuracy,
they did not succeed with the economics of power and bill of materials for portable 
goods. Stereo cameras cost less, but need a lot of computational processing, and this 
aspect is getting good attention of research community , spurring the development  of 
FPGA
and GPU based acceleration of stereo-vision related computation. The low power 
consumption of fpga-based solutions are attractive and crucial for high performance 
embedded computing too.

  
This paper describes our design and implementation of a 
real-time stereo depth estimation system with Zedboard\cite{zedboard} 
(housing ARM-SoC based FPGA) at its center. 
This system uses Zed stereo camera\cite{zedcamera}, Intel RealSense D435i\cite{intel} or ELP stereo-camera for capturing images. 
Real-time Raster-Respecting Semi-Global Matching\cite{r3sgm} (R3SGM) along with Census Transform are used for disparity estimation. 
The system takes in real-time data from the cameras and generates a depth image from it. 
Rectification of the images, as well as stereo matching, is implemented in the FPGA 
whereas capturing data from USB cameras and controlling the FPGA peripherals is done via 
application programs which run on the hard ARM processor on Zedboard. Development of the 
FPGA IP's is done using High-Level Synthesis (HLS) tools. A VGA monitor is interfaced to 
Zedboard to display the computed depth image in real-time.

Our approach is inspired by R3SGM\cite{r3sgm} a hardware implementation of SGM. Table 
\ref{fpga_comparision} ( at the later portion of the paper ) shows the comparison of hardware utilization between our approach 
and \cite{r3sgm} which shows ours uses much lesser Hardware Resources and thus having less 
power consumption. It may be emphasized that we have focused on very low power consumption as well as small form factor that is
necessary for drones vision, blind aid etc.


\section{Literature Review}
\par There has been a lot of research on the topic of disparity map generation dating back to 1980s. \cite{lit_rev} reviews most of the works including both software and hardware implementations. \par A binocular Stereo Camera estimates disparity or the difference in the position of the pixel of a corresponding location in the camera view by finding similarities in the left and right image. There have been various costs governing the extent of the similarity. Some of them are Sum of Absolute Differences(SAD), Sum of Squared Differences(SSD), Normalized Cross-Correlation and the recent Rank Transform and Census Transforms. They are window-based local approaches where the cost value of a particular window in the left image is compared to the right image window by spanning it along a horizontal axis for multiple disparity ranges. The window coordinate for which the metric cost is the least is selected which gives us the disparity for that corresponding center pixel. 
From the disparity, the depth value is computed by equation \ref{depth_equation} where 
the baseline is the distance between the optical centers of two cameras. 
\begin{equation}
    Depth = Baseline * (Focal Length)/ disparity
    \label{depth_equation}
\end{equation}
Local window-based approaches suffer when the matching is not reliable which mostly happens 
when there are very few features in the surrounding. This results in the rapid variations
of the disparities. This problem is solved by global approaches which use a smoothing 
cost to penalize wide variations in the disparity and trying to propagate the cost across
various pixels. The following are some of the global approaches. 

\subsection{Semi Global Matching (SGM)}
\par SGM is a stereo disparity estimation method based on global cost function 
minimization.  Various versions of this method (SGM, SGBM, SGBM forest) are still among 
the top-performing stereo algorithm on Middlebury datasets. This method minimizes the 
global cost function between the base image and match image and a smoothness constraint 
that penalizes sudden changes in neighboring disparities. Mutual information between 
images, which is defined as the negative of joint entropy of the two images, is used in 
the paper\cite{sgm} as a distance metric. Other distance metrics can also be used with a 
similar effect as has been demonstrated with census distance metric in our 
implementation. Since we already had a Census Implementation, we used it for our SGM 
implementation. The Hamming Distance returned by Census stereo matching is used as the 
matching cost function for SGM. The parameters for Census are window size 7x7, disparity 
search range 92. The image resolution is 640x480. Sum of Absolute 
Differences (SAD) was also considered as a matching cost function. But it was observed 
that SAD implementation consumes more FPGA resources than the Census implementation with 
same parameters. This may be due to the fact that SAD computation is an arithmetic 
operation whereas Census computation is a logical operation.  

\par Simple census stereo matching has a cost computation step in which for a particular 
pixel we generate an array of costs (Hamming distances). The length of this array is 
equal to the disparity search range. The next step is cost minimization in which the 
minimum of this array (minimum cost) is computed and the index of the minimum cost is 
assigned as disparity. In SGM, an additional step of cost aggregation is performed 
between cost computation and cost minimization. The aggregated cost for a particular 
pixel p for a disparity index d is given by equation \ref{sgm_eq}.

\begin{equation}
    \begin{split}
L_{r}(p,d) = C(p.d)+ min(L_{r}(p-r,d),\\L_{r}(p-r,d-1)+P_{1},\\L_{r}(p-r,d+1)+P_{1},\\min_{i}(L_{r}(p-r,i)+P_{2}))\\
-min_{k}(L_{r}(p-r,k))
    \label{sgm_eq}
    \end{split}
\end{equation}
For each pixel at direction ‘r’, the aggregated cost is computed by adding the current 
cost and minimum of the previous pixel cost by taking care of penalties as shown in 
Equation \ref{sgm_eq}. First-term $C(p,d)$ is the pixel matching cost for disparity $d$. 
In our case, it is the Hamming distance returned by Census window matching. It is 
apparent 
that the algorithm is recursive in the sense that to find the aggregated cost of a pixel 
$L'_{r}(p, )$, one requires the aggregated cost of its neighbors $L'_{r}(p-r,)$. $P_{1}$ 
and $P_{2}$ are empirically determined constants. For detailed discussion refer to \cite{sgm}. 

\subsection{More Global Matching (MGM)}
\par As SGM tries to minimize the cost along a line it suffers from streaking effect. 
When there is texture less surface or plane surface the matching function of census 
vector may return different values in two adjacent rows but due to SGM, the wrong 
disparity may get propagated along one of the paths and can 
result in streaking lines.

\par
\begin{figure}[h]
    \centering
    \subfloat[MGM in General]{
    \includegraphics[width=0.4\linewidth]{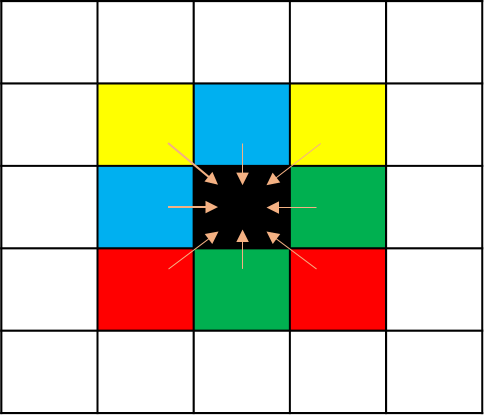}}
    \hspace{5pt}
    \subfloat[Our Implementation]{
    \includegraphics[width=0.4\linewidth]{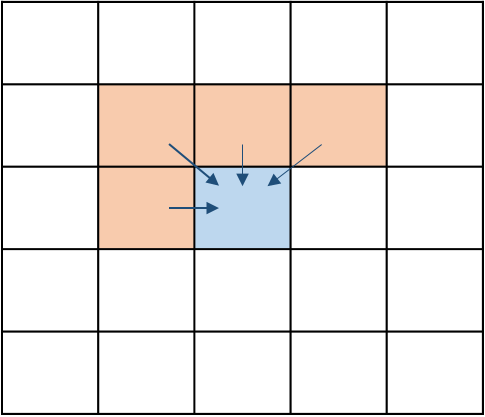}}
    \caption{Grouping of Paths in MGM}
    \label{mgm_path_figure}
\end{figure}

 MGM\cite{mgm} solves this problem by taking the average of the path cost along 2 or 
more paths incorporating information from multiple paths into a single cost. It uses this 
result for the next pixel in the recursion of Equation \ref{sgm_eq}. The resultant 
aggregated cost at a pixel is then given by the Equation \ref{mgm_eq}  
\begin{equation}
    \begin{split}
L_{r}(p,d) = C(p.d)+1/n\sum_{x\varepsilon{\{r_n\}}} (min(L_{r}(p-x,d),\\L_{r}(p-x,d-1)+P_{1},\\L_{r}(p-x,d+1)+P_{1},\\min_{i}(L_{r}(p-x,i)+P_{2}))\\
-min_{k}(L_{r}(p-x,k)))
    \label{mgm_eq}
    \end{split}
\end{equation}
where n has the value depending on the number of paths that we want to integrate into the
information of single cost. For example, in Figure \ref{mgm_path_figure}$a$ two paths are
grouped into 1 so $n$ has value 2 and there are a total of 4 groups. Thus we need to 
store 4 cost vectors in this case and while updating 1 cost value in the center pixel 
have to read cost vector of the same group from 2 pixels. Lets say $r=1$ for blue boxes 
group in Figure \ref{mgm_path_figure}$a$, while updating the $L_r$ for this group of the 
centre pixel in Equation \ref{mgm_eq} we have $x$ as left and top pixels. From here on 
SGM refers to MGM variant of it.

\section{Hardware architecture and Implementation}
\subsection{System Design}
\par
\begin{figure}[h]
  \includegraphics[width=1\linewidth]{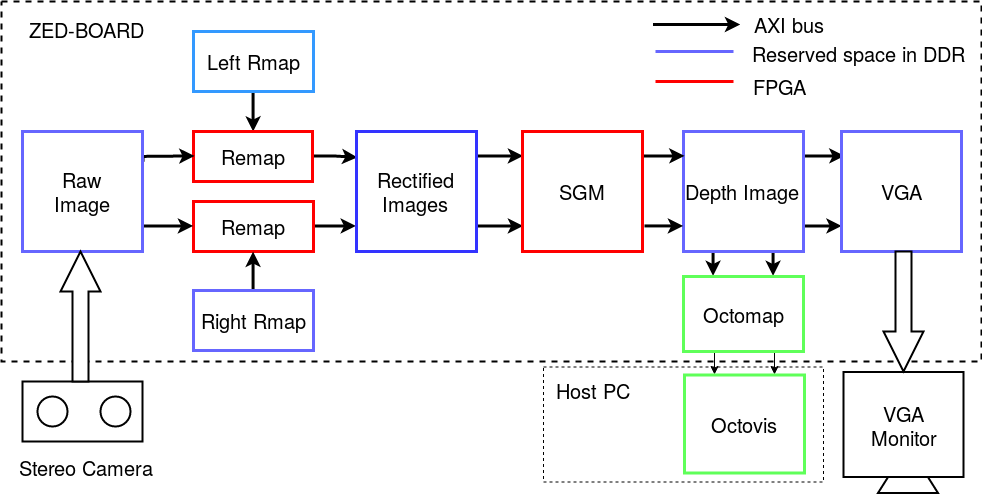}
  \centering
  \caption{Block diagram}
  \label{main_block_diagram}
\end{figure}

 Figure \ref{main_block_diagram} shows an overview of the implemented system. Left 
and right images captured from the Zed camera\cite{zedcamera} are stored into DDR RAM 
(off-chip RAM). Maps required for the stereo rectification of the images are statically 
generated offline using OpenCV\cite{opencv}. These maps are also stored into DDR RAM. We 
need two Remap peripherals which perform stereo rectification for the left and right 
images respectively. The Remap peripheral reads the raw image frame and the corresponding
map and generates a rectified image frame. The rectified images are again stored into 
DDR. The Intel RealSense camera requires USB3.0 or higher to stream left and right 
images. However, Zedboard does not have USB3.0. Hence the camera cannot be directly 
interfaced to the board. So images were continuously captured and streamed from a 
computer using ethernet. The left and right image streams were received by a socket 
client running on the ARM processor on Zedboard. The camera outputs rectified images, 
hence remap peripheral is not required in this case. The images received from the socket 
client are stored into DDR RAM. We have also implemented it for Zed Camera 
\cite{zedcamera}. For both camera modules in Binocular cameras, the stereo matching 
peripheral (SGM block in the figure\ref{main_block_diagram}) then reads the left and 
right rectified frame and generates disparity image which is again stored into DDR. The 
VGA peripheral is configured to read and display the disparity image onto a VGA monitor. 
FPGA peripherals perform memory access using the AXI4 protocol.
\par The resolution of images is fixed to 640x480 and cameras are configured accordingly.
Each pixel is stored as an eight-bit number. The metric used to profile the computation 
times of different peripherals and also the cameras is fps (frames per second). From here on a frame means 640x480 pixels.
\par We could have skipped storing the rectified images and passed the output of the 
Remap peripheral directly to the stereo matching peripheral. We chose not to do this 
because our performance is not limited by memory read-write but by the FPGA peripherals
themselves. We use the AXI4 protocol to perform memory read-write. The read-write rates 
are 3 orders of magnitude greater than the compute times of FPGA peripherals.

\par The images are captured using application programs running on the ARM processor on 
Zedboard. The programs make use of v4l2 library for image capture. The ARM processor is 
also used to control the FPGA peripherals. 

\subsection{Undistortion and Rectification}
\par
\begin{figure}[h]
  \includegraphics[width=0.8\linewidth]{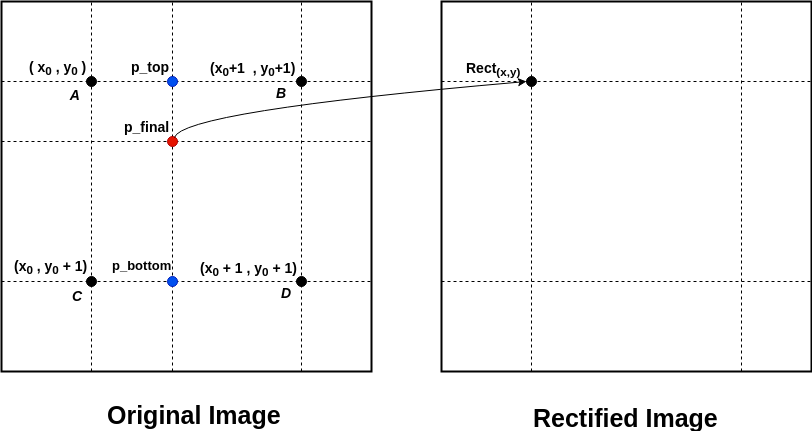}
  \centering
  \caption{Remap operation}
  \label{remap_interp}
\end{figure}

 Stereo camera calibration and rectification (one time step) is done using the OpenCV
library. Calibration and rectification process produces distortion coefficients and 
camera matrix. From these parameters, using the OpenCV library, two maps are generated, 
one for each camera. Size of a map is the same as image size. Rectified images are built 
by picking up pixel values from raw images as dictated by the maps. The map entry (i,j) 
contains a coordinate pair (x, y); and the (i, j) pixel in the rectified image gets the 
value of the pixel at (x, y) from the raw image. x and y values need not be integers. In 
such a case, linear interpolation is used to produce final pixel value. Figure 
\ref{remap_interp} shows the remap operation with 4 neighbour bilinear interpolation.
\par On-chip memory is limited in size, and it is required by the stereo-depth hardware 
module. So, we store the maps generated during calibration and rectification in system
DDR. The map entries are in fixed-point format with five fractional bits. Captured images
are stored in DDR too. The hardware module iterates over the maps, and builds up the 
result (left and right) images by picking pixels from raw images. Note that, while the 
maps can be read in a streaming manner, the random-access is required for reading the raw
images. For fractional map values, bilinear interpolation (fixed point) is performed. 
Resulting images are stored back in DDR. As this hardware module has to only - "read maps
and raw images pixels from DDR, perform bilinear interpolation, and store the pixels 
back", it needs less than 5\% resources of the Zynq chip.
\subsection{SGM Block Architecture}
\par
\begin{figure}[h]
    \centering
    \includegraphics[width=0.4\linewidth]{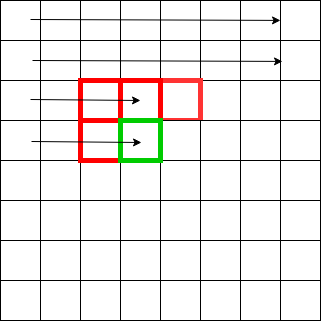}
    \caption{Four neighbour paths considered for SGM}
    \label{fig:sgm_4_paths}
\end{figure}

\begin{figure}[h]
    \centering
    \includegraphics[width=\linewidth]{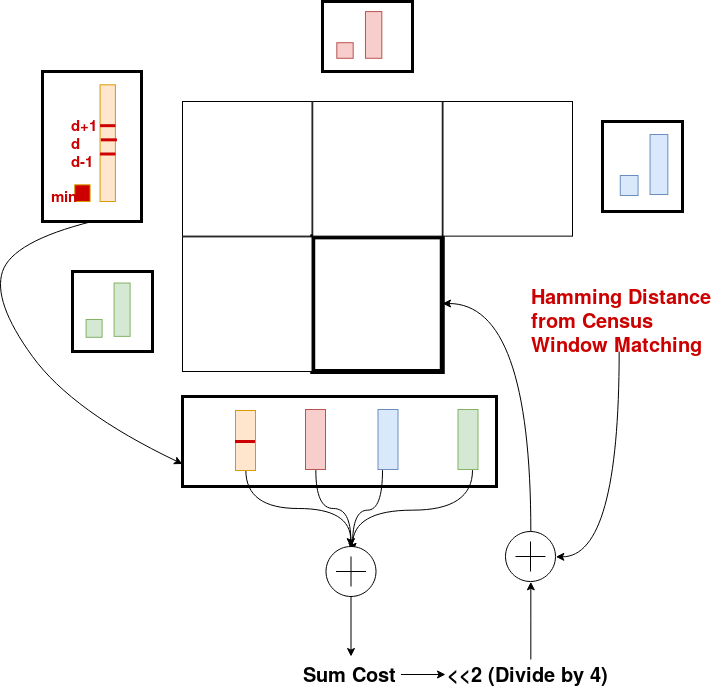}
    \caption{SGM Cost Computation. Steps involved in calculating the disparity for the current pixel.}
    \label{fig:sgm_sum_compute}
\end{figure}

 In Census implementation we scan using row-major order through every pixel in the 
image and perform stereo matching. Thus for the SGM implementation built upon this, we 
consider only four neighbors for a pixel under processing as shown in red in Figure
\ref{fig:sgm_4_paths}. This is done because we have the required data from neighbors
along these paths. The quality degradation by using 4 paths instead of 8 paths is 
2-4\%\cite{sgm_4paths_quality}.
Figure \ref{fig:sgm_sum_compute} shows the implemented SGM architecture. The aggregated 
cost for all paths and disparity indices of one row above the pixel (full row not shown 
in figure) and the left adjacent pixel of the current pixel are depicted as columns of
colour yellow, red, blue and green for paths top left, top, top right and left 
respectively.We store the resultant accumulated cost which is computed using Equation 
\ref{mgm_eq}. 4 Paths have been used by grouping them into single information as shown in
Figure \ref{mgm_path_figure}b. Thus in Equation \ref{mgm_eq} our n value in 4 and r has a
single value for a pixel. The Census metric cost is stored in an 8bit unsigned char so 
the total size of memory occupied by the cost is given as $Size of Row Cost Array = 
(Image Width)*(Disparity Range)*(No of Path Groups) = 640*92*1 = 57.5KB $.

Minimum cost across disparity search range is computed once and stored for the above row 
and left adjacent pixel. These scalar quantities are shown as small boxes of the same 
color. Since the minimum cost values are accessed multiple times, storing the minimum 
values instead of recomputing them every time they are required saves a lot of 
computations. The pixels in the row above the current pixel can be either top-left, top 
or top-right neighbors of the current pixels. Hence costs along the left path (green 
columns) are not stored for the row above the pixel.  
Figure \ref{fig:sgm_sum_compute} also shows the data required and the steps for computing
the aggregated cost for a certain pixel considering all the 4 paths. Smoothing term(2nd 
part in the RHS of Equation \ref{mgm_eq}) along all paths are summed up to obtain a sum 
cost which has to be divided by n(4). Since division is resource-intensive hardware we 
use left a shift by 2 to divide by 4. Then the resulting value is added with the current 
hamming distance (1st part in the RHS of Equation \ref{mgm_eq}). An upper bound is 
applied to the sum cost. The index of the minimum of this modified sum cost is the 
disparity for this pixel. The costs for all disparities are stored as they will be 
required for future pixels of the next row. The minimum cost across the disparity search 
range is also computed and stored for all paths.

\par
\begin{figure}
    \centering
    \includegraphics[width=1\linewidth]{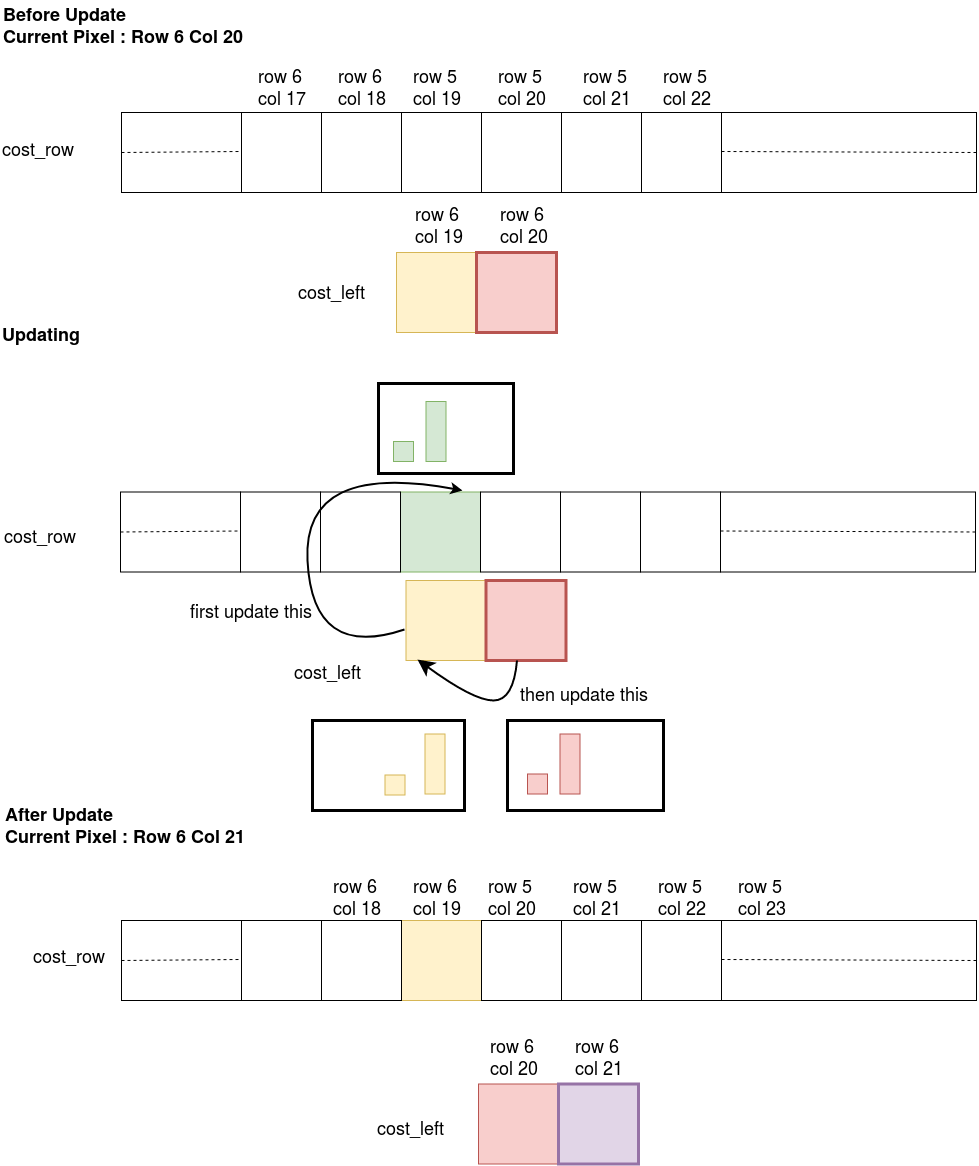}
    \caption{SGM Array Updation. }
    \label{fig:sgm_arr_upd}
\end{figure}

 Figure \ref{fig:sgm_arr_upd} shows the data structures used for storing the costs 
and the algorithm for updating them as we iterate over pixels. The $cost\_row$ structure 
has dimensions- image columns, path groups and disparity search range. It stores the 
costs for one row above the current pixel for all paths and disparity indices. The 
$cost\_left$ structure has dimensions- path groups and disparity search range. It stores 
the cost for the left adjacent pixel of the current pixel for all paths and disparity 
indices. As shown in Figure \ref{fig:sgm_arr_upd} the current pixel under processing is 
at row 6 column 20. It requires data from its 4 neighbors: row 5 column 19, row 5 column 
20, row 5 column 21 and row 6 column 19.To generate data for current pixel we use the 
data of $cost\_left$ and 3 pixel vectors of $cost\_row$. As we compute the disparity for 
this pixel and also performing the housekeeping tasks of generating the required data, we
update the structures as shown in Figure \ref{fig:sgm_arr_upd}. The data from 
$cost\_left$ is moved to the top-left neighbour of the current pixel in $cost\_row$. The 
top left pixel cost data is not required anymore and hence is not stored. After this 
update is done, the currently generated data is moved into $cost\_left$.

\par Pixels at the top, left and right edge of the image are considered to have neighbors
with a maximum value of aggregated cost. As SGM cost aggregation step is a minimization 
function, they are effectively ignored. The $cost\_row$ and $cost\_left$ structures are 
initialized to a maximum value before the stereo matching process. This initialization 
has to be done for every frame.

\subsection{HLS Implementation}
High-level Synthesis(HLS) platform such as Vivado HLS ( from Xilinx ) facilitates 
a suitably annotated description of compute-architecture in high level language like
C or C++ , which it converts to a low-level HDL based description of the same
computing architecture. The generated VHDL or Verilog code is then synthesized
to target fpgas. 
We have used Vivado HLS tools provided by 
Xilinx to convert our C implementation to HDL and 
package it to an IP for further use. 
The structure of HLS stereo matching code is as follows.
\begin{lstlisting}
void stereo_matching_function(){
for(int row=0; row<IMG_HEIGHT; row++) {
 for(int col=0; col<IMG_WIDTH; col++) {
  //Reading pixel from DDR through AXI4 
  protocol in row-major order
  //Shifting the Census Match window in 
  the left and right blocks
  for(int d=0; d<SEARCH_RANGE; d++) {
   //Match l_window with r_window[d]
   //Update the min cost index 
   //Add the necessary output to the cost 
   row and cost left vectors
  }
 //write disparity image pixel to DDR
 }
} 
}
\end{lstlisting}
\par There are no operations between the row and col loop, hence they can be effectively 
flattened into a single loop. The plan was to pipeline the merged row-column loop. Thus 
resulting in increase of frame rate by disparity range times if the pipeline throughput 
had been 1. However the resources in fpga device on Zedboard are not enough to permit the pipelining 
the row column loop. Hence, only the search range loop was pipelined. The arrays used in 
the implementation have been partitioned effectively to reduce the latency.
Based on the availability of Hardware resources we have divided the whole image into 
sections and disparity of each section is computed in parallel.
It was observed that a frame rate of $2.1$ fps is obtained with the most used resource 
being Block RAM (BRAM) 17\%.  The time required for processing one frame for such an 
implementation can be given as
\begin{equation}
\begin{split}
    T \propto no.\; of\; rows\; \times no.\; of\; columns\; \times \\ (search\; range\; + pipeline\; depth\;)
\label{time_req}    
\end{split}
\end{equation}
\par The characteristic of this implementation is that the logic synthesized roughly 
corresponds to the matching of two Census windows, the cost aggregation arithmetic and 
on-chip memory to store data for the next iterations. As we sequentially iterate over 
rows, columns and disparity search range we reuse the same hardware. Thus, the FPGA 
resources required are independent of the number of rows, columns and search range but 
computation time required is proportional to these parameters as shown by equation 
\ref{time_req}. This gives us the idea to divide the images into a number of sections 
along the rows and process the sections independently by multiple such SGM blocks. As the
most used resource is BRAM at 17\%, we can fit 5 such SGM blocks with each block having 
to process 5 sections of the image i.e. 128 rows in parallel. Thus we increase resource 
usage 5 times and reduced the time required for computation by the same resulting in $10.5$ fps.

\par
\begin{figure}[h]
    \centering
    \includegraphics[width=1\linewidth,height = 0.6\linewidth]{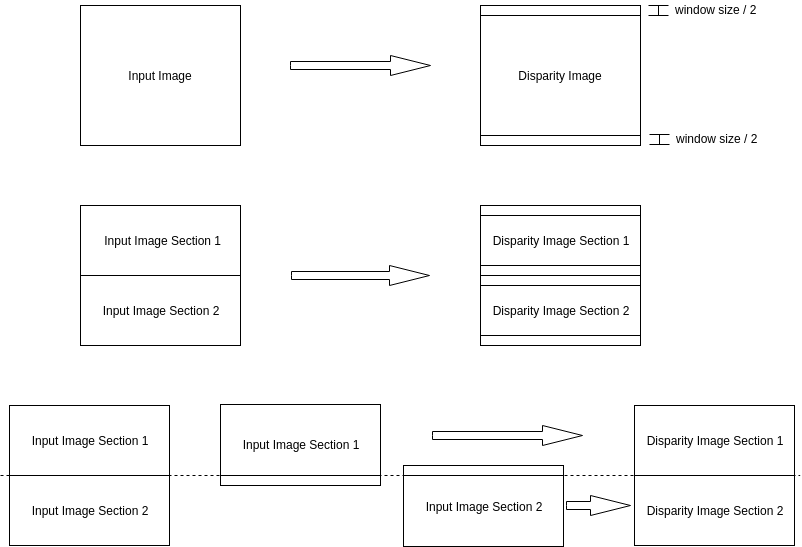}
    \caption{Dividing the input image into two sections to be processed by two blocks simultaneously}
    \label{fig:multiblocks}
\end{figure}

 One flaw to this approach is that if we divide the input image into exactly 5 parts,
there will be a strip of width window size at the center of the disparity image where the
pixels will be invalid. The solution to this is that the height of each section is 
$image\_height/5 + window\_size/2$. This is shown in Figure \ref{fig:multiblocks} for an 
example of 2 sections. 

\subsection{Hardware Setup}
\par
\begin{figure}[h]
    \centering
    \includegraphics[width=1\linewidth]{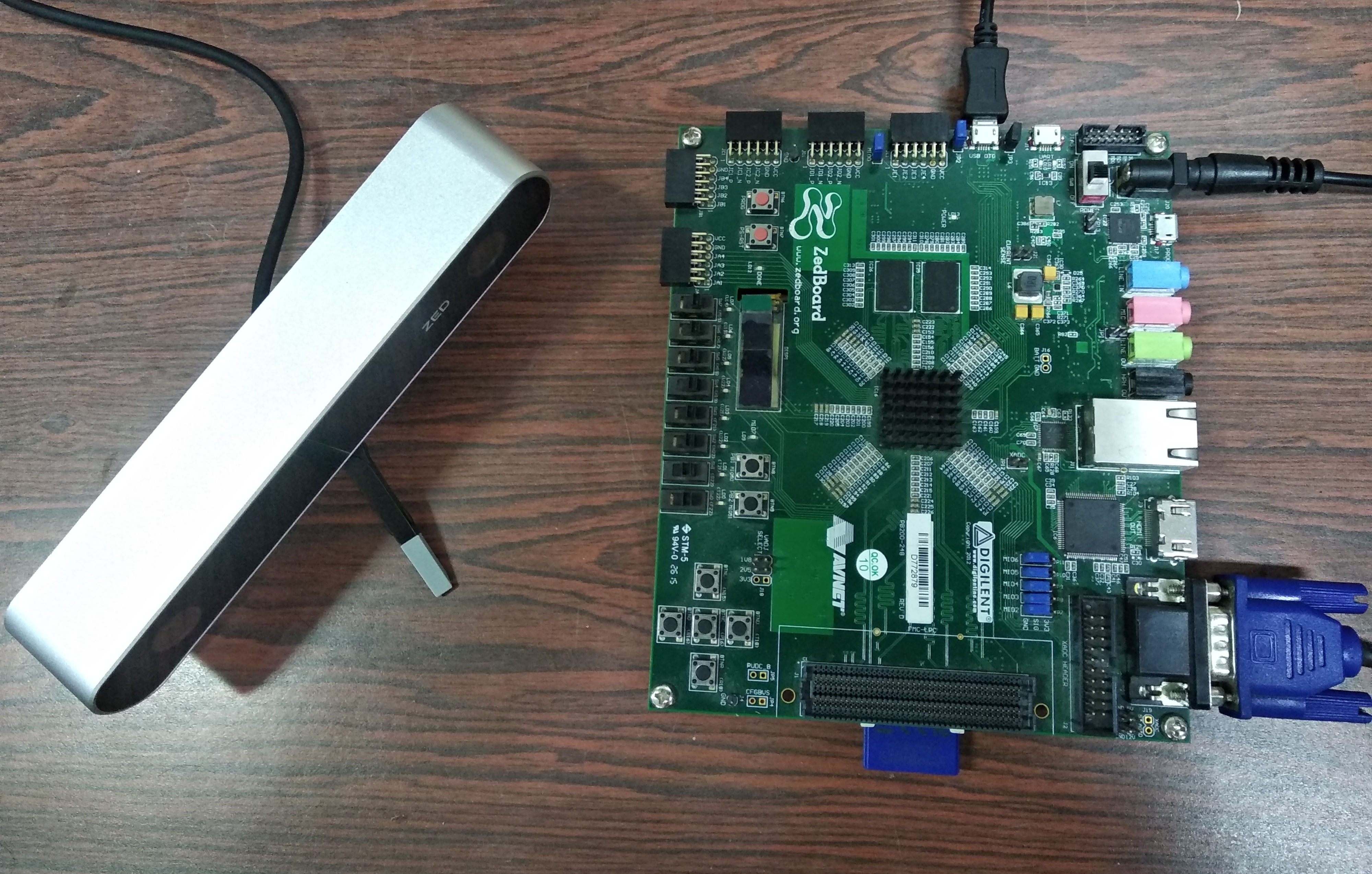}
    \caption{Hardware setup}
    \label{fig:hardware_setup}
\end{figure}

 Figure \ref{fig:hardware_setup} shows the hardware setup. The Zed camera is 
connected to a USB 2.0 port of the Zedboard. The Zedboard is booted with petalinux 
through SD card. In the case where Intel RealSense camera is used, we require ethernet to
receive the images. The only other connections to Zedboard are the connection to VGA 
display and power. 

\section{Experimental Results and evaluation}
\par The obtained frame rate for the implemented system is $10.5$ fps with Zedboard running
at 100 MHz. The Power consumption of the computation which is performed in FPGA is 0.72W 
whereas the on-chip arm processor which is being used to capture the images and start the
FPGA peripherals along with the ELP stereo-camera  consumes 1.68 watt ,
thereby raising consumption to 2.4W. A $10m\Omega $, 1W current sense resistor is in series with the 12V input power supply on the Zedboard. 
Header J21 straddles this resistor to measure the voltage across this resistor 
for calculating Zedboard power\cite{zedboard}. The 
resource usage is summarized in Table \ref{res_util}. 
It is observed that the BRAM  utilization is the most. This is due to storing large cost arrays.
\begin{table}[h]
\begin{tabular}{|c|c|c|c|c|c|} 
    \hline
    & BRAM & DSP & FF & LUT & LUTRAM \\ [0.5ex] 
    \hline
    Utilization &  132 & 65 & 39159 & 37070 & 981 \\ 
    \hline
    Available & 140 & 220 & 106400 & 53200 & 17400\\
    \hline
    \% Utilization & 94.3 & 29.5 & 36.8 & 69.6 & 5.64\\
    \hline
\end{tabular}
\caption{\label{res_util}Resource utilization for the entire design in Zedboard}
\end{table}
\begin{figure}[h]
\centering
\subfloat[Left image]{\includegraphics[width=1.6in]{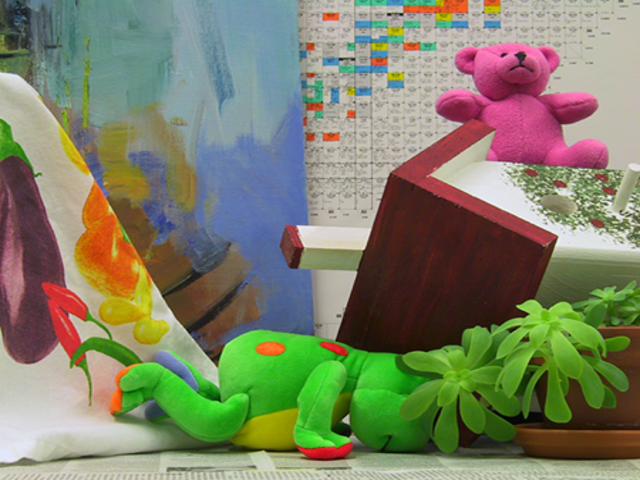}}
\hspace{10pt}
\subfloat[Ground truth]{\includegraphics[width=1.6in]{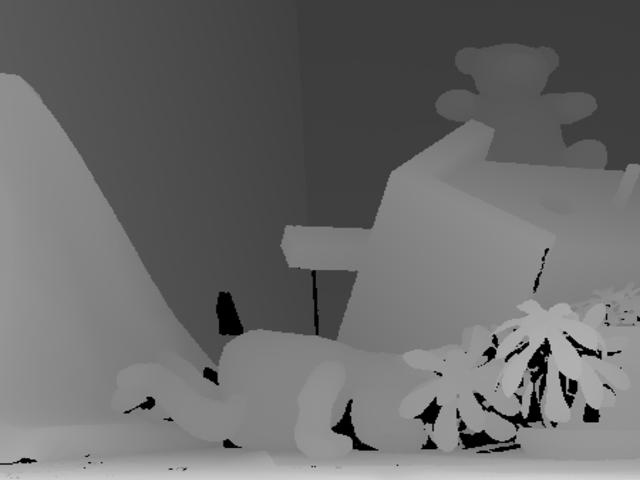}}
\\
\subfloat[SGM 4 paths software]{\includegraphics[width=1.6in]{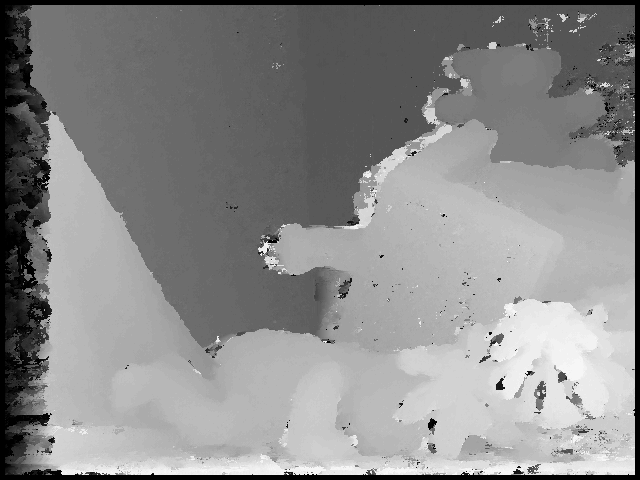}}
\hspace{10pt}
\subfloat[SGM 8 paths software]{\includegraphics[width=1.6in]{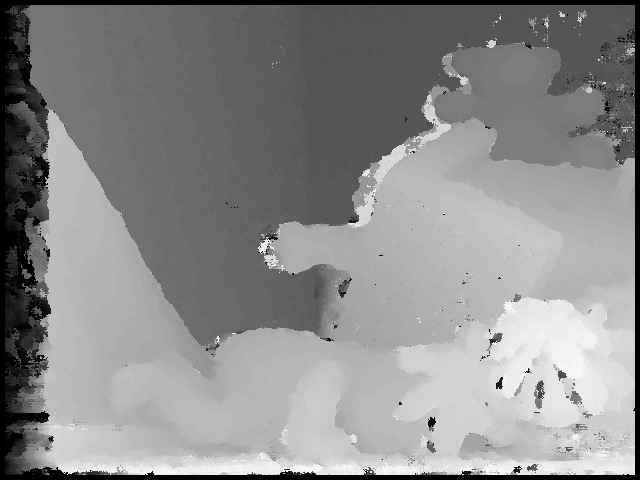}}
\\
\subfloat[SGM 4 paths hardware]{\includegraphics[width=1.6in]{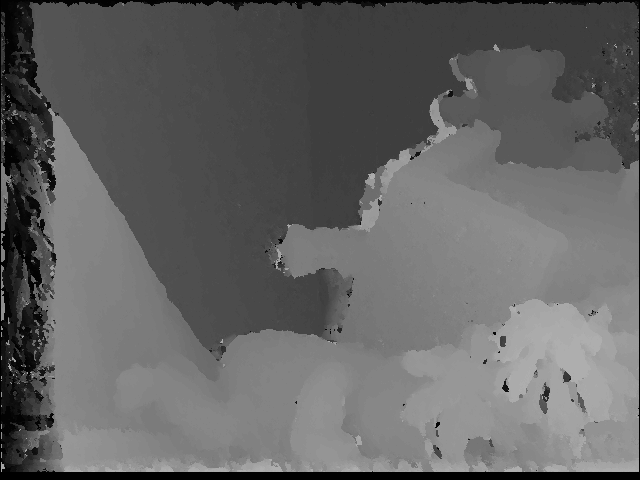}}
\hspace{10pt}
\subfloat[SGM with arrays initialized to zeros]{\includegraphics[width=1.6in]{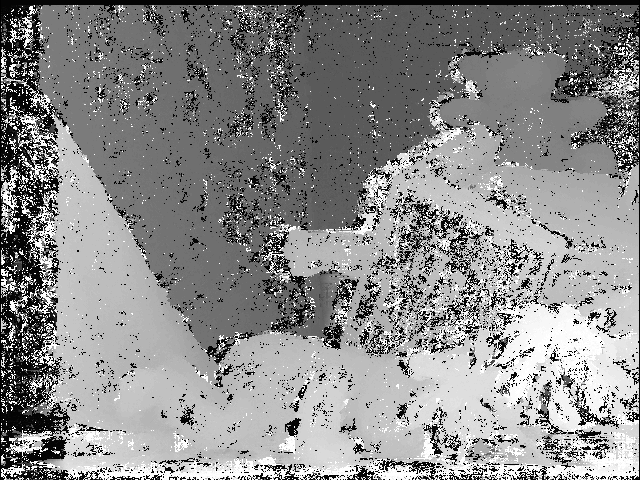}}
\caption{SGM results on Middlebury images }
\label{fig:sgm_mberry_output}
\end{figure}
\par The algorithmic accuracy is measured using Root mean square of difference in the disparity values obtained by our implementation with the ground truth on Middlebury test images given in Table 
\ref{percent_error} column 2. It can also be measured by percentages of erroneous disparities in Table 
\ref{percent_error} column 3. A 5 pixel tolerance is 
considered due to intensity variation caused by changing resolution of raw image. It is notable that no post processing has been done on the SGM 
output.
\begin{table}
\centering
\begin{tabular}{|c|c|c|} 
     \hline
     Image  & RMSE & \% Erroneous disparities \\ [0.5ex] 
     \hline
     Teddy & 5.43 & 11\\ 
     \hline
     Dolls & 6.79 & 17 \\
     \hline
     Books & 6.82 & 20\\
     \hline
     Moebius & 7.54 & 20\\
     \hline
     Laundry & 9.22 & 27 \\
     \hline
     Reindeer & 9.17 & 27 \\
     \hline
     Art & 9.24 & 30 \\
     \hline
\end{tabular}
\caption{\label{percent_error}Accuracy metric of ours disparity image pixels as compared to 
ground truth for Middlebury images}
\end{table}

\begin{figure}[h]
    \centering
    \subfloat[Left image with IR blaster on]{\includegraphics[width=1.6in]{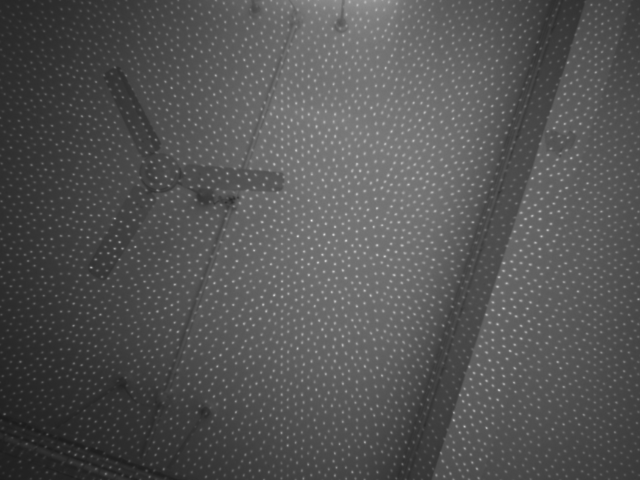}}
    \hspace{10pt}
    \subfloat[Left image with IR blaster covered]{\includegraphics[width=1.6in]{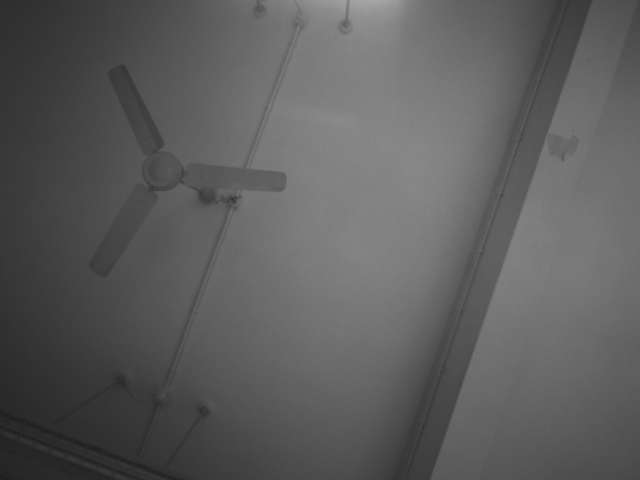}}
    \\
    \subfloat[Disparity image from camera with blaster on]{\includegraphics[width=1.6in]{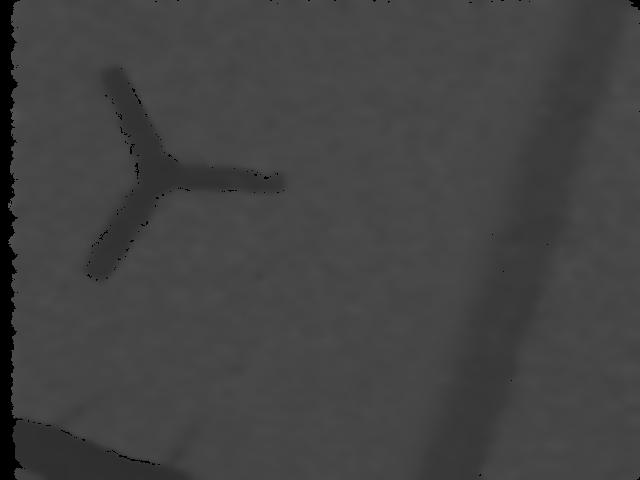}}
    \hspace{10pt}
    \subfloat[Disparity image from camera with IR blaster covered]{\includegraphics[width=1.6in]{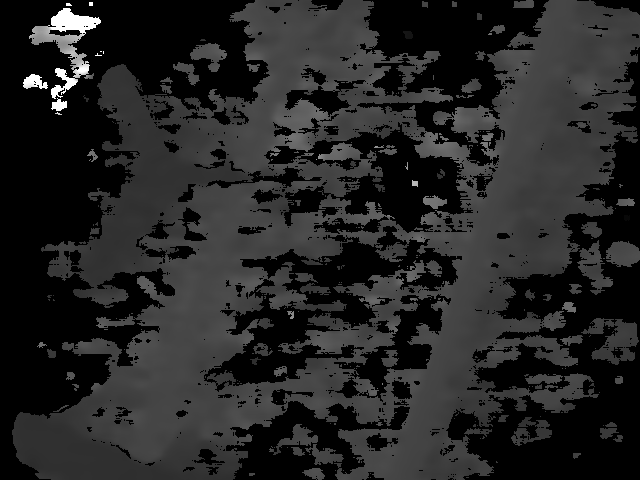}}
    \\
    \subfloat[SGM disparity image with blaster on]{\includegraphics[width=1.6in]{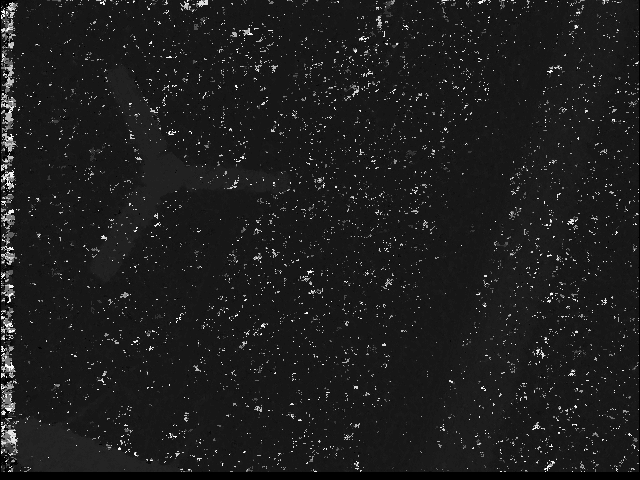}}
    \hspace{10pt}
    \subfloat[SGM hardware disparity image with IR blaster covered]{\includegraphics[width=1.6in]{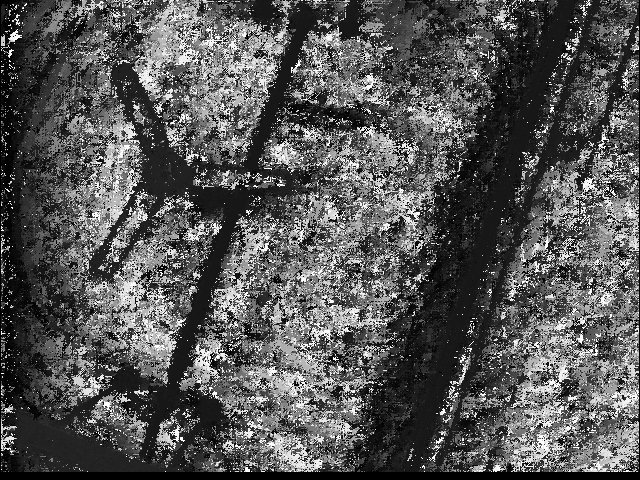}}
    \caption{SGM results on Realsense image: effect of texture }
    \label{fig:sgm_rs_texture}
\end{figure}

\begin{figure*}
    \centering
    \subfloat{\includegraphics[width=1.6in]{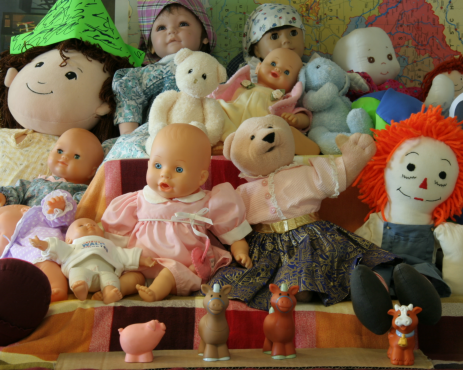}}
    \hspace{3pt}
    \subfloat{\includegraphics[width=1.6in]{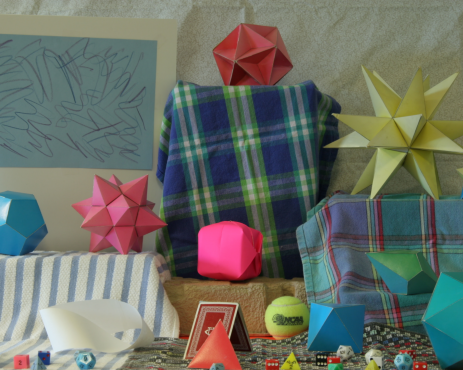}}
    \hspace{3pt}
    \subfloat{\includegraphics[width=1.6in]{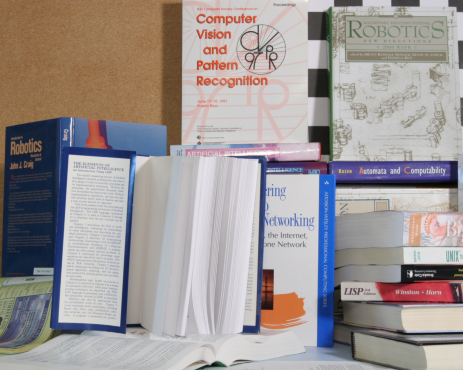}}
    \hspace{3pt}
    \subfloat{\includegraphics[width=1.6in]{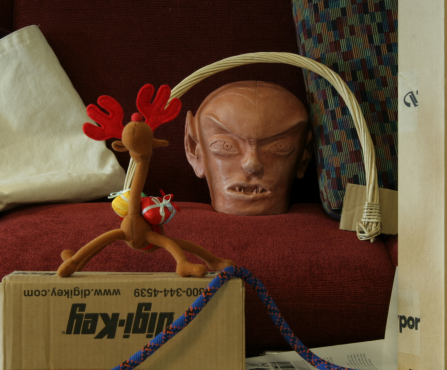}}
    \\[0.3ex] 
    \subfloat{\includegraphics[width=1.6in]{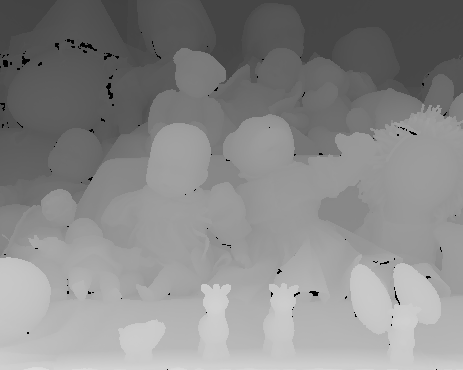}}
    \hspace{3pt}
    \subfloat{\includegraphics[width=1.6in]{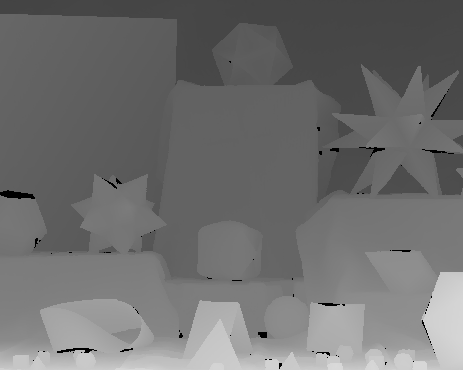}}
    \hspace{3pt}
    \subfloat{\includegraphics[width=1.6in]{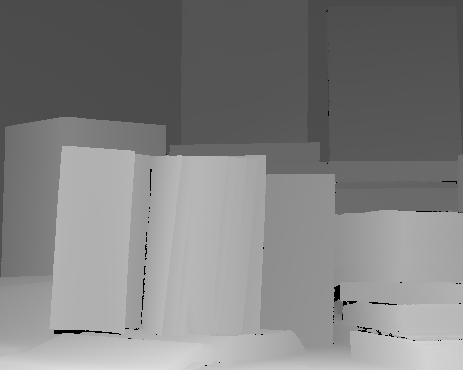}}
    \hspace{3pt}
    \subfloat{\includegraphics[width=1.6in]{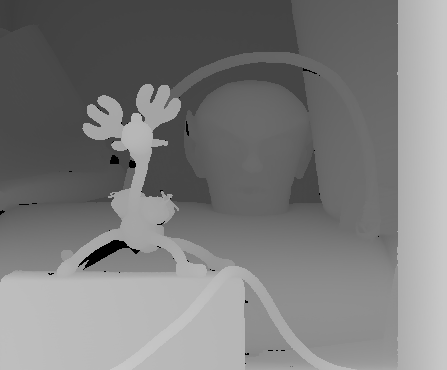}}
    \\[0.3ex]
    \subfloat{\includegraphics[width=1.6in]{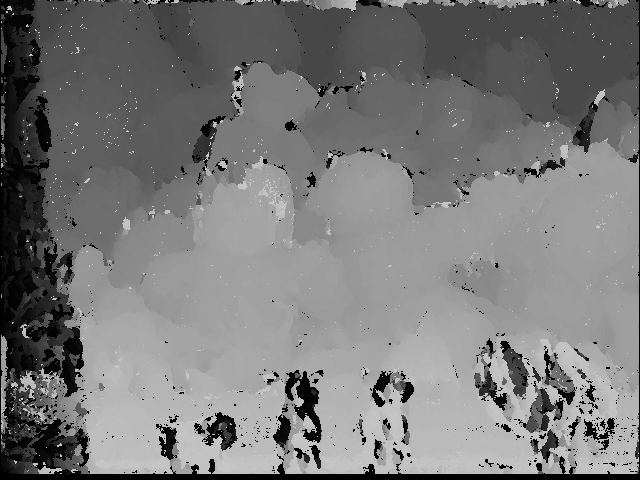}}
    \hspace{3pt}
    \subfloat{\includegraphics[width=1.6in]{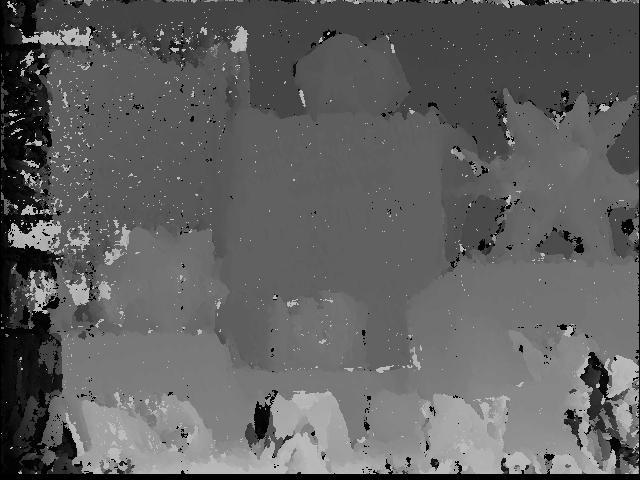}}
    \hspace{3pt}
    \subfloat{\includegraphics[width=1.6in]{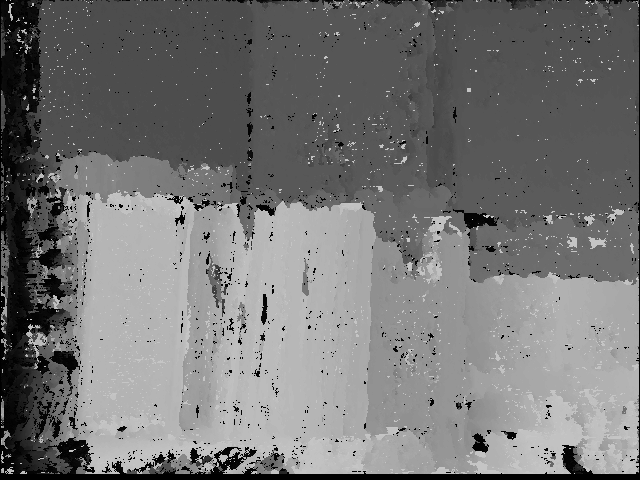}}
    \hspace{3pt}
    \subfloat{\includegraphics[width=1.6in]{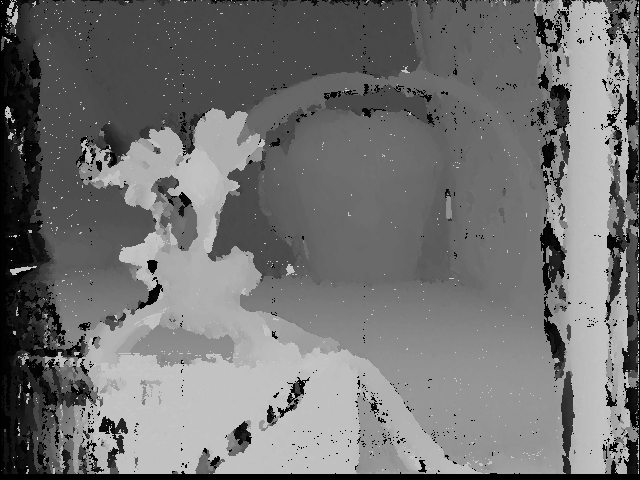}}
    \caption{Qualitative Comparison of our results with some of the Middlebury data set. 1st Row contains the Left Raw Images, 2nd Row contains the ground truth of the corresponding Images and 3rd Row contains the Output of our Implementation. }
    \label{mgm_results}
\end{figure*}
\par Figure \ref{fig:sgm_mberry_output} shows the software and hardware implementation 
results on Teddy image from Middlebury 2003 dataset\cite{middlebury}. Figure 
\ref{fig:sgm_mberry_output}c-d show the results of an inhouse software implementation of 
SGM and Figure  \ref{fig:sgm_mberry_output}e shows the result of the hardware 
implementation. It can be observed that SGM with 8 paths gives the best results. SGM with
4 paths in software gives slightly better results than the hardware implementation. The 
difference in results is due to the fact that the way the algorithm is implemented in 
software and hardware is different. Figure \ref{fig:sgm_mberry_output}f shows the SGM 
disparity image with $cost\_row$ and $cost\_left$ initialized to zero. Since the cost 
aggregation function is minimization function, the zeros from the arrays propagate to 
further pixels. The trickle down effect causes the degradation of the disparity image. 
Similar results with frame rate around 8.3 fps were also achieved by an inhouse GPU implementation of SGM on Jetson TK1 board which is of MAXWELL architecture with 256 cores and power consumption $< 10$ watts. This implementation is analyzed and optimized by using
OpenMP for multi-threading and AVX (Advanced Vector Extension) registers for vectorization. GPU shared memory is used to reduce the global memory access. CUDA shuffle instructions are used to speed-up the algorithm and vector processing is also applied.

\par
\begin{figure}[h]
\centering
\subfloat[Left image classroom]{\includegraphics[width=1.6in,height=1.2in]{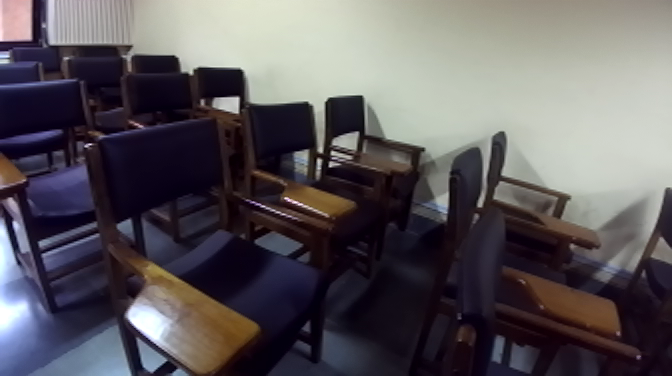}}
\hspace{10pt}
\subfloat[SGM disparity image 
classroom]{\includegraphics[width=1.6in,height=1.2in]{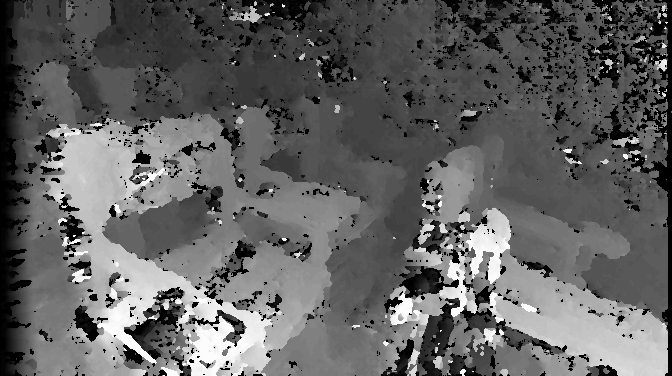}}
\caption{SGM results on ZED camera image}
\label{fig:sgm_zed}
\end{figure}

\begin{figure}[h]
    \centering
    \subfloat[Left image]{\includegraphics[width=1.6in]{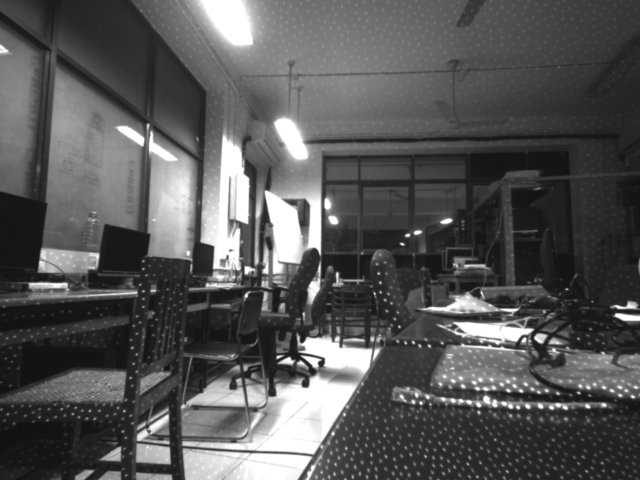}}
    \hspace{10pt}
    \subfloat[disparity image from the camera]{\includegraphics[width=1.6in]{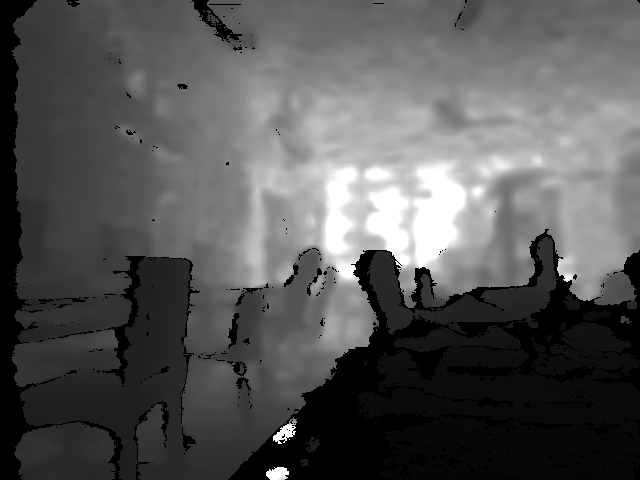}}
    \hspace{10pt}
    \subfloat[SGM hardware disparity image]{\includegraphics[width=1.6in]{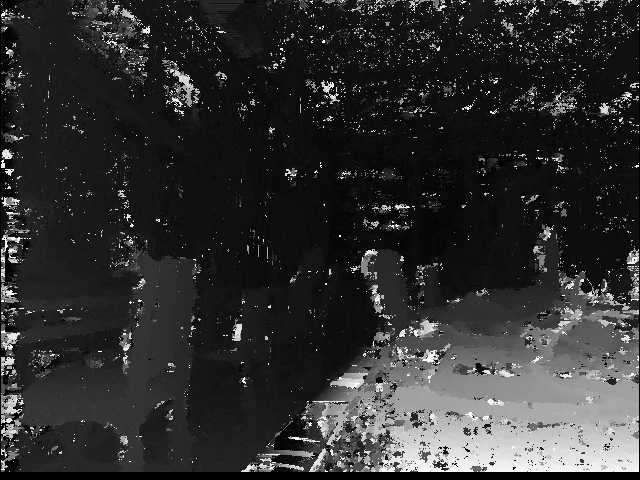}}
    \caption{SGM results on Realsense image: lab }
    \label{fig:sgm_rs_output_lab}
\end{figure}

 Fig \ref{fig:sgm_zed} and \ref{fig:sgm_rs_output_lab} shows the captured image and 
the corresponding disparity image obtained using the SGM implementation. The Intel 
RealSense camera also provides a disparity image. This is shown in Figure \ref{fig:sgm_rs_output_lab}b. The convention followed here is opposite i.e closer objects
appear darker. 

\par The Intel RealSense camera has an infrared (IR) light projector which projects 
structured light onto the scene. This pattern can be seen in Figure 
\ref{fig:sgm_rs_output_lab}a. Figure \ref{fig:sgm_rs_texture} shows the effect of the 
infrared projector on disparity estimation. Figure \ref{fig:sgm_rs_texture}ace show the 
captured left image from the camera, disparity image obtained from the camera and the 
computed disparity image when IR blaster was on. Figure \ref{fig:sgm_rs_texture}bdf show 
the same images when the IR blaster was covered. Incase of \ref{fig:sgm_rs_texture}e 
although the image contains salt noise, it can be easily filtered out. The fan blades can
be easily seen in the disparity image. In \ref{fig:sgm_rs_texture}f there are more number
of white pixels which imply that the object is very near to the camera which is a false 
result. As can be seen, the structured light projector helps in stereo matching by adding
texture to non-textured surfaces.

\par
\begin{figure}[h]
    \centering
    \includegraphics[width=1\linewidth]{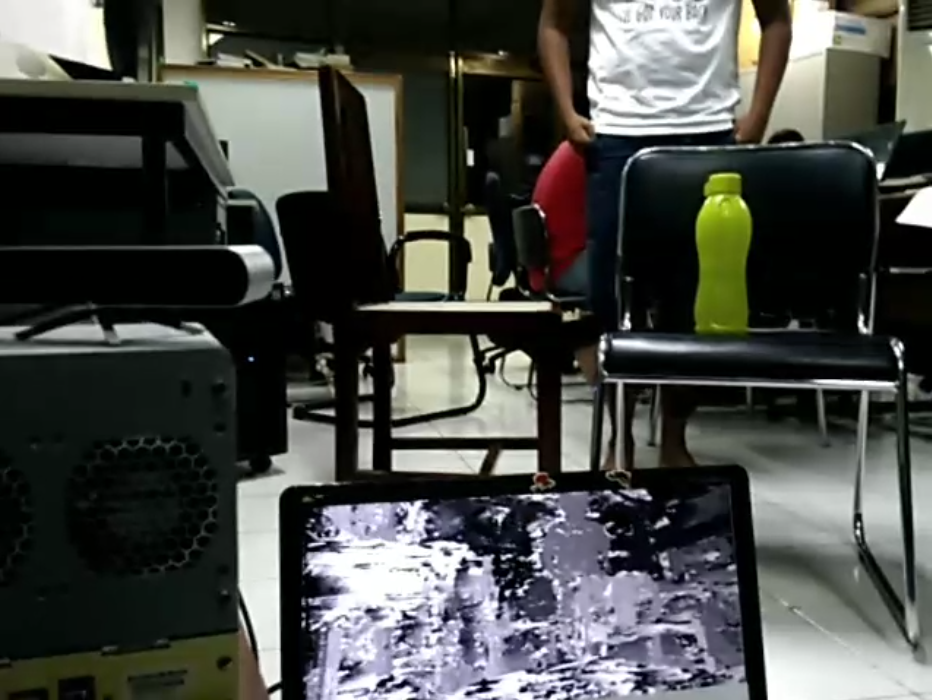}
    \caption{Scene and disparity image on VGA monitor}
    \label{fig:video_output}
\end{figure}

 Figure \ref{fig:video_output} shows the scene and the corresponding disparity image 
obtained on the VGA monitor. The camera can be seen on the left side of the image.

\par Figure \ref{mgm_results} shows the qualitative comparison or our results with 
Middlebury data set. We can see that the objects placed near are not accurate this is 
because we have used the disparity range of 92 pixels and so it is not able to find a 
match in the corresponding left and right images. Thus for a better accuracy, disparity range can be
increased with the trade-off being update rate as the pipeline latency will increase.

Finally we inform the reader about our comparison with R3SGM \cite{r3sgm} work.
Table 
\ref{fpga_comparision} shows the comparison of hardware utilization between our approach 
and \cite{r3sgm} which shows ours uses much lesser Hardware Resources and thus having less 
power consumption. Furthermore, if we were to use fpga used in \cite{r3sgm}, 
we would have far more liberty with resources that can be leveraged to 
further pipeline the design and obtain another order of speedup. However we have focused on very low power consumption as well as small form factor that is
necessary for drones vision, blind aid etc.
We can extrapolate the frame rate likely to be achieved by our design on
ZC706 board as below.
We can replicate the hardware four times (assuming other resources are
under limit) to utilize all of the BRAM,
and get 40fps performance.
However, it would increase the power consumed by zynq chip, as well as
by camera and DDR subsystems for this higher frame capture and 
processing rate.

\begin{table}[h]
\centering
\begin{tabular}{|c|c|c|c|c|p{0.8cm}|p{1cm}|} 
    \hline
    & BRAM18K & DSP & FF & LUT & Frame Rate & Power (Approx) \\ [0.5ex] 
    \hline
    Ours & 132 & 65 & 39159 & 37070 & 10.5&  0.72W\\ 
    \hline
    \cite{r3sgm} & 163 & - & 153000 & 109300 & 72 & 3W\\
    \hline
\end{tabular}
\caption{\label{fpga_comparision}Comparison of FPGA Hardware Resources(Approx) and power consumption between our approach and \cite{r3sgm}}
\end{table}

\section{Conclusion}
In this paper we presented the hardware implementation of the MGM\cite{mgm} which is a 
variant of SGM\cite{sgm} on Zedboard\cite{zedboard} an FPGA-ARM based SOC inspired by 
R3SGM\cite{r3sgm}. In order to reduce the memory consumption, we have grouped 4 paths- 
left, top left, top, and top right, whose pixel data are available while processing as a 
result of row-major order streaming process. The efficient utilization of hardware 
resources resulted in a low power consumption of 0.72W for data processing on FPGA that 
computes the Rectification and disparity Map generation and with 1.68W for data acquisition from 
Cameras along with starting the peripherals using the on board ARM processor achieving an
update rate of 10.5Hz with a good accuracy as was shown in Table\ref{percent_error} and 
Figure\ref{mgm_results}. 
This system is highly suitable to be used in micro UAVs, blind Aids or any portable types
of equipment with a small form factor and high power constraints.

\end{document}